# Satellite Image Utilization for Dehazing with Swin Transformer-Hybrid U-Net and Watershed loss

Jongwook Si[a], Sungyoung Kim[b]

[a]Dept. of Computer · AI Convergence Engineering, Kumoh National Institute of Technology, Gumi, Republic of Korea
[b]School of Computer Engineering, Kumoh National Institute of Technology, Gumi, Republic of Korea

**Abstract**

Satellite imagery plays a crucial role in various fields; however, atmospheric interference and haze significantly degrade image clarity and reduce the accuracy of information extraction. To address these challenges, this paper proposes a hybrid dehazing framework that integrates Swin Transformer and U-Net to balance global context learning and local detail restoration, called SUFERNOBWA. The proposed network employs SwinRRDB, a Swin Transformer-based Residual-in-Residual Dense Block, in both the encoder and decoder to effectively extract features. This module enables the joint learning of global contextual information and fine spatial structures, which is crucial for structural preservation in satellite image. Furthermore, we introduce a composite loss function that combines L2 loss, guided loss, and a novel watershed loss, which enhances structural boundary preservation and ensures pixel-level accuracy. This architecture enables robust dehazing under diverse atmospheric conditions while maintaining structural consistency across restored images. Experimental results demonstrate that the proposed method outperforms state-of-the-art models on both the RICE and SateHaze1K datasets. Specifically, on the RICE dataset, the proposed approach achieved a PSNR of 33.24 dB and an SSIM of 0.967, which is a significant improvement over existing method. This study provides an effective solution for mitigating atmospheric interference in satellite imagery and highlights its potential applicability across diverse remote sensing applications.

"Keywords:Dehazing; Satellite Image; Watershed; Transformer"

## 1. Introduction

Satellite imagery has been demonstrated to play a crucial role in a variety of engineering applications, including environmental monitoring, climate change analysis, agricultural management, national security, map transformation, and disaster response [1-4]. With recent advancements in artificial intelligence (AI) and high-resolution sensor technology, automated satellite image analysis systems have become increasingly sophisticated, contributing significantly to diverse industries and research domains [5-8]. However, atmospheric interference is a significant challenge, as it degrades the clarity of satellite images and consequently reduces the accuracy of the extracted information [9-11]. Among the major sources of atmospheric interference, clouds and fog/haze are particularly significant. In Low Earth Orbit (LEO) satellite images, water vapor and aerosols in the atmosphere cause light scattering, leading to reduced image contrast and localized color distortion [12-13]. Furthermore, if atmospheric interference is not adequately removed, critical surface information may be lost, which has deleterious effects on applications such as object detection and change detection.

The utilization of satellite imagery has become pervasive across diverse disciplines, with environmental monitoring being a pivotal application. However, should satellite images be degraded due to atmospheric interference, the accuracy of such analyses may be compromised, potentially leading to erroneous interpretations. Consequently, the development of image dehazing techniques that can effectively eliminate atmospheric interference [14] and restore clear imagery is imperative for enhancing the reliability of environmental monitoring [15]. Furthermore, disaster response and management heavily rely on satellite imagery [16]. In situations such as floods, earthquakes and hurricanes, the ability to rapidly assess damage and formulate recovery plans is essential. Satellite imagery is a powerful tool for large-scale disaster monitoring, but clouds and haze often obscure critical regions, making it difficult to obtain a clear view of affected areas. To address this issue, image dehazing techniques must be applied to enhance image quality and ensure accurate disaster assessment. Satellite imagery has also been extensively adopted in the domains of agriculture and resource management [17-18]. It plays a pivotal role in monitoring crop health, predicting yields, and detecting droughts or pest infestations. However, when atmospheric interference distorts satellite imagery, the reliability of agricultural assessments is significantly compromised. This, in turn, can have a detrimental effect on the decision-making process in agriculture. The development of robust dehazing techniques is therefore essential for ensuring the accurate interpretation of satellite imagery.

Atmospheric degradation in satellite imagery primarily manifests in the form of haze and clouds, both of which are major causes of reduced image quality. Haze results from the scattering of light by fine particles in the atmosphere, such as aerosols or water vapor, and typically appears as a thin, uniform layer that lowers image contrast and color fidelity across the entire scene. In contrast, clouds are dense and spatially non-uniform occlusions that can completely obscure structural and textural information. Moreover, cloud boundaries often exhibit semi-transparency and cast shadows, further complicating the restoration process. Although haze and clouds differ in their physical characteristics, they both represent atmospheric degradations and can therefore be addressed within a unified dehazing framework. However, haze removal requires global contrast restoration capabilities, whereas cloud removal demands localized, nonlinear reconstruction to recover occluded regions effectively.

Given the extensive range of applications in various industries and research domains, satellite image dehazing is imperative for ensuring the accurate extraction and analysis of information [19-21]. Nevertheless, existing image restoration techniques encounter numerous limitations, particularly due to the absence of robust dehazing models specifically designed for satellite imagery. Consequently, there is an urgent requirement to establish a novel methodology that can effectively eliminate atmospheric interference while maintaining the critical image features.

Existing studies on satellite image dehazing can largely be categorized into two groups: physics-based models and deep learning-based models [22-41]. Physics-based models try to remove atmospheric interference by using atmospheric scattering models, which estimate atmospheric light and transmission maps to reconstruct the original image. These methods are predicated on well-defined physical principles and can be effective in specific environments where atmospheric conditions are relatively stable. However, they are encumbered by significant limitations that render them impractical for large-scale applications. A key challenge confronting these models is their inability to generalize effectively across diverse atmospheric conditions, resulting in variable and often unsatisfactory performance. Furthermore, the requirement for auxiliary information, such as depth maps or multi-spectral data, which may not always be available, represents a significant drawback. Furthermore, the complexity inherent in accurately modelling real-world atmospheric effects frequently results in incomplete or inaccurate dehazing, especially in highly dynamic environments where atmospheric properties fluctuate.

In contrast, deep learning-based models leverage Convolutional Neural Networks (CNNs) and Transformer architectures to learn dehazing patterns from large-scale datasets. These models have demonstrated impressive performance by automatically extracting complex features and adapting to different atmospheric conditions. Nevertheless, they are also characterized by a number of limitations. One key challenge is that their performance is heavily dependent on the quality of the training set and is susceptible to domain bias. This can lead to the model's ability to perform well on certain datasets but to fail to generalize to unseen situations. Furthermore, many deep learning-based models introduce spatial distortions, which affect the structural information of objects within the image. Additionally, these models frequently apply smoothing techniques that result in the loss of fine detail and essential texture information, which is particularly problematic for high-resolution satellite imagery where fine-grained detail is critical for analysis.

The distinctive properties of satellite imagery further complicate the direct application of existing dehazing techniques. In contrast to ground-based images, satellite images capture vast geographic areas, frequently under highly variable atmospheric conditions. As a consequence, conventional dehazing models, which were originally designed for natural scene images, tend to introduce spatial deformations, alter color distributions when applied to satellite data. It is therefore evident that a robust dehazing framework tailored specifically for satellite imagery is necessary to achieve high-quality restoration while preserving essential scene information.

In order to address the challenges posed by atmospheric interference in satellite imagery, this study proposes the **SUFERNOBWA** network (**S**atellite image **U**tilization **F**or d**E**hazing with swin t**R**a**N**sf**O**rmer-hy**B**rid u-net and **WA**tershed loss). This study proposes a hybrid network architecture that integrates the Swin Transformer with the U-Net framework. The Swin Transformer employs a Shifted Window-based Multi-head Self-Attention mechanism, which enables efficient modeling of long-range dependencies and global contextual features within the image. This architectural design is particularly advantageous for satellite imagery, where various structural elements are distributed over large spatial extents, and haze effects span across broad regions. The ability to extract hierarchical global features ensures semantic consistency and effective dehazing over wide areas. However, Transformer-based models are generally less effective at preserving local fine-grained details, such as boundaries and textures. To address this limitation, the Swin Transformer modules are embedded within a U-Net architecture, which follows an encoder–decoder design. U-Net is capable of multi-scale feature extraction and leverages skip connections to directly pass high-resolution spatial information from the encoder to the decoder. This mechanism is particularly effective in maintaining local structural integrity, including object edges, textures, and shapes, which are critical in high-resolution satellite imagery.

As a result, the proposed hybrid architecture simultaneously leverages the global feature modeling capability of the Swin Transformer and the local detail preservation strength of U-Net, effectively addressing the unique challenges of satellite image dehazing— namely, global haze modeling across large spatial domains and accurate reconstruction of local structures. Its primary objectives are to eliminate atmospheric interference, to preserve the structural integrity of satellite imagery, and to optimize the quality of image restoration. The study's primary contributions are as follows:

- **A novel dehazing network combining Swin Transformer [42] and U-Net**

  Unlike conventional CNN-based networks that focus primarily on local information, this study integrates Swin Transformer for global feature extraction and U-Net for multi-scale feature learning, ensuring an effective dehazing architecture tailored for satellite imagery.

- **A new loss function based on the watershed algorithm**

  The watershed algorithm, which has historically been employed for object segmentation, is integrated into the loss function with the objective of preserving object structures while effectively removing atmospheric interference.

- **Optimization using a combination of guided loss and watershed loss**

  The proposed method integrates guided loss to preserve boundary details and enhance structural consistency in dehazed images.

The remainder of this paper is organized as follows. Section 2 comprises a review of related research on existing dehazing techniques and satellite image restoration. Section 3 presents the proposed SUFERNOBWA model and loss function design. Section 4 contains experimental outcomes and performance evaluations, which are then compared with the results obtained from current approaches. Finally, Section 5 offers a summary of the paper's findings, a discussion of its results, and an outline of future research directions.

## 2. Related Works

The domain of image dehazing has undergone extensive research and development, establishing itself as a critical area in computer vision and image processing due to its significance in enhancing visual clarity and information extraction under degraded atmospheric conditions. B. Huang et al. [22] proposed a dehazing method utilizing SAR and RGB images. By applying a conditional generative adversarial network and dilated residual blocks, they effectively fused RGB and SAR information to achieve more efficient haze removal. This study demonstrated superior performance over existing models by leveraging multi-sensor data, enabling more accurate restoration of hazy remote sensing images. The key difference from this study is the adoption of deep learning architecture, which is designed to effectively learn global contextual information using only RGB images. X. Chen et al. [23] proposed a hybrid high-resolution learning network to restore fine spatial details in satellite images affected by atmospheric interference. This model ensures spatially precise outputs using a high-resolution branch while effectively integrating multi-resolution information through multi-resolution convolution streams and the Parallel Cross-Scale Fusion module. Additionally, the Channel Feature Refinement block is applied to adjust features dynamically across channels, enhancing dehazing performance. A commonality with the proposed method is the utilization of multi-scale information fusion for atmospheric interference removal. However, the key difference lies in the approach: while H2RL-Net relies on high-resolution learning and multi-resolution convolutions, SUFERNOBWA employs a U-Net structure designed to effectively capture global contextual information.

T. Song et al. proposed the Transformer-based RSDformer [24] model for remote sensing image dehazing. RSDformer incorporates three key modules to effectively learn both global and local dependencies. It employs the Detail-Compensated Transposed Attention mechanism to extract both global and local features across channels, which allows the model to handle non-uniformly distributed haze more effectively. Compared to the proposed SUFERNOBWA model, both approaches leverage Transformer architectures, highlighting their commonality in capturing long-range dependencies. However, while RSDformer relies on Transformer-based attention mechanisms to directly learn long-range dependencies, SUFERNOBWA integrates Swin Transformer into a hybrid U-Net framework. L. Yang et al. proposed DehFormer [25], a Transformer-based model designed to enhance dehazing performance for high-resolution satellite images. DehFormer integrates three key modules: Multi-Scale Feature Aggregation Network for multi-resolution feature fusion, Gated-Dconv Feed-Forward Network for efficient feature selection, and Multi-Dconv Head Transposed Attention to optimize Transformer computations by reducing matrix multiplications and leveraging frequency domain processing. These design improvements enable DehFormer to outperform state-of-the-art dehazing methods. Compared to SUFERNOBWA, both models employ Transformer-based architectures and aim to capture both global and local features for improved restoration. However, SUFERNOBWA introduces a Watershed-based loss function to enhance boundary preservation, while DehFormer emphasizes multi-resolution feature fusion and frequency-adaptive processing for improved dehazing performance.

H. Hwang et al. [26] utilizes CycleGAN to perform translation between Synthetic Aperture Radar images and optical images. Their approach enables mutual conversion between SAR and optical images by employing two generator networks and two discriminator networks trained simultaneously. Compared to SUFERNOBWA, it focuses on transforming SAR and optical images, emphasizing style transfer and structural information preservation by incorporating SSIM (Structural Similarity Index Measure) and Perceptual Loss into the loss function. In contrast, SUFERNOBWA employs a Swin Transformer-based hybrid U-Net structure to remove atmospheric interference within a single optical image while incorporating Watershed loss to preserve boundary information, highlighting a key distinction between the two approaches. P. Singh et al. proposed Cloud-GAN [27], a method for cloud removal in remote sensing (RS) images. Cloud-GAN operates by learning the mapping between cloudy and cloud-free images, utilizing an adversarial loss to ensure that the generated de-clouded images closely match the distribution of real cloud-free images. Additionally, a cycle consistency loss is incorporated to enforce that the generated cloud-free images retain the same scene as the original cloudy images. Compared to SUFERNOBWA, Cloud-GAN is a GAN-based model primarily focused on cloud removal, whereas SUFERNOBWA is designed to remove atmospheric interference. Furthermore, while Cloud-GAN employs cycle consistency loss to maintain scene consistency, SUFERNOBWA integrates Watershed loss to enhance boundary preservation, highlighting a key distinction between the two approaches.

B. Cai et al. proposed DehazeNet [28], a CNN-based approach for single-image dehazing. DehazeNet estimates the medium transmission map from the input hazy image and applies it to the atmospheric scattering model to restore a clear image. It utilizes Maxout units for feature extraction and introduces a novel activation function called Bilateral ReLU to enhance the dehazing performance. Both DehazeNet and SUFERNOBWA adopt CNN-based approaches for haze removal. However, while DehazeNet relies solely on a CNN-based architecture to learn physical constraints, SUFERNOBWA integrates both CNN and Transformer structures to achieve more effective feature learning. S. Li et al. designed M2SCN [29] to enhance dehazing performance by incorporating a Multi-Model Joint Estimation module and a Self-Correcting module. The M2JE module formulates the dehazing process as a multi-model ensemble problem to improve the generalization capability of the model. Meanwhile, the SC module gradually corrects errors in the intermediate features extracted by the network, allowing it to effectively process images with non-homogeneous haze. Compared to the proposed SUFERNOBWA, both models adopt an end-to-end learning framework for remote sensing image dehazing, aiming for more effective restoration of hazy images. However, M2SCN focuses on improving generalization performance and correcting intermediate features to address non-uniform haze degradation. In contrast, SUFERNOBWA applies Watershed loss to preserve edge information, ensuring more accurate haze removal, which distinguishes it from M2SCN.

S. Vishwakarma et al. [30] proposed a machine learning-based approach to effectively remove atmospheric interference in satellite images. First, in the preprocessing stage, techniques such as histogram equalization and color adjustment are applied to enhance the initial image quality. Next, the dehazing network utilizes CNNs, residual blocks, attention mechanisms, and GANs to remove atmospheric interference and restore image quality. Finally, in the postprocessing stage, detail enhancement, noise removal, and color correction are performed to produce natural and sharp image results. Compared to SUFERNOBWA, this study focuses on performing dehazing and further improving the final image quality through additional postprocessing steps. HALP [31] developed a novel dehazing algorithm to address the issue of quality degradation in remote sensing images captured in hazy weather. The proposed method introduces a dehazing technique based on Heterogeneous Atmospheric Light estimation and the Side Window Filter. Compared to SUFERNOBWA, HALP adopts a physics-based dehazing approach by estimating non-uniform atmospheric light in local regions and improving the filtering process. In contrast, SUFERNOBWA aims to enhance dehazing performance through a deep learning-based approach and the integration of loss functions.

Y. Guo et al. proposed SCANet [32] to effectively address the issue of non-homogeneous haze. This network is designed with an attention generator network and a scene reconstruction network, focusing on enhancing regions severely affected by haze. Compared to SUFERNOBWA, both models share common goals, such as haze removal, fine detail restoration, and enhancement of haze-occluded regions. Additionally, both models adopt a hybrid architecture that integrates local and global feature extraction methods. M. Wang et al. proposed IDF-CR [33], a novel cloud removal method that leverages the powerful generative capabilities of the diffusion model. IDF-CR adopts a two-stage approach that combines pixel space processing and iterative noise diffusion in latent space to progressively enhance restoration performance. Compared to the proposed SUFERNOBWA, both models share the commonality of utilizing advanced generative models to improve image restoration performance. However, while IDF-CR employs a diffusion model to iteratively remove clouds, SUFERNOBWA removes haze through a reconstruction-based approach.

SpAGAN [34] is a model that removes clouds from remote sensing imagery using a GAN-based spatial attention mechanism. It mimics the human visual attention mechanism to identify cloud regions and enhance restoration performance by focusing on information from a local-to-global perspective.

Similar to SUFERNOBWA, it employs a neural network-based restoration approach and selectively processes cloud regions to improve image quality. However, while SpAGAN applies GAN training with a spatial attention mechanism at its core, the proposed method differentiates itself by employing multiple loss functions and specialized network blocks to enhance structural information and edge preservation. DehazeFormer [35] proposes a novel approach by utilizing the Swin Transformer. However, it identifies that the original design of the Swin Transformer is not well-suited for image dehazing and improves its performance by modifying the normalization scheme, activation function, and other components. Both SUFERNOBWA and DehazeFormer share the common goal of enhancing dehazing performance using a Swin Transformer-based architecture. However, while DehazeFormer focuses on addressing the structural limitations of the Swin Transformer, the proposed method prioritizes preserving structural information, such as buildings and roads.

Lihe et al. [36] proposed PhDnet, a physics-aware dehazing network for remote sensing images that integrates a haze extraction unit based on the atmospheric scattering model. To enhance multi-scale feature fusion, the network adopts Multi-Scale Gating Convolution (MSGC). While both PhDnet and SUFERNOBWA utilize CNN-based architectures for haze removal, PhDnet emphasizes physical interpretability by incorporating domain knowledge, whereas SUFERNOBWA focuses on learning global contextual representations using a hybrid Swin Transformer and U-Net framework. Yang et al. [37] introduced a Dual-View Knowledge Transfer (DVKT) framework, which leverages a teacher-student paradigm to transfer knowledge from natural image dehazing models to remote sensing scenarios. The use of intra-layer and inter-layer knowledge transfer modules enables effective feature learning in lightweight models. Unlike SUFERNOBWA, which prioritizes structural integrity and boundary-aware restoration via a dedicated loss function, DVKT focuses on model compression and efficiency through knowledge distillation. Despite both models leveraging multi-scale feature processing, their design objectives differ significantly. C. Li et al. [38] proposed an efficient dehazing method applicable to both outdoor and remote sensing images, utilizing Gaussian-weighted image fusion to improve the accuracy of transmittance estimation and employing an unsharp mask-based approach for correcting color distortion. Similar to our study, this method shares the objective of improving transmittance accuracy while preserving color consistency; however, it primarily relies on traditional restoration and enhancement techniques, which limit its capability in feature representation and learning-based restoration inherent to deep learning approaches.

In UAVD-Net [39], a dehazing framework specifically designed to remove spatially non-uniform haze in UAV remote sensing images was proposed, combining a Transformer-based global information capturing module with a local information enhancement module to perform global-local feature collaborative dehazing. Our method similarly utilizes Swin Transformer-based global information along with local restoration, yet differs in that we employ SwinRRDB to achieve a lighter and more efficient structure, and we additionally incorporate a watershed-based structure-preserving loss tailored to the characteristics of satellite images. Furthermore, the UDAVM-Net [40] study introduced a U-shaped Dual Attention Vision Mamba Network, applying dual attention mechanisms and Residual Vision Mamba Blocks (RVMBs) within a U-Net structure for single-image dehazing in satellite remote sensing. This study leveraged the global effective receptive field (ERF) capabilities of the Mamba architecture as a Transformer alternative and utilized an atmospheric radiative transfer model and cloud distortion model for dataset construction. While it shares the use of a U-Net-based multi-scale structure for leveraging both global and local features similar to our work, our method differentiates itself by employing window-based attention mechanisms in the Swin Transformer for improved computational efficiency and by incorporating a structure-preserving loss design. The ICL-Net [14] study utilized an Inverse Cognitive Learning Network to perform dehazing through multi-scale feature extraction and adaptive learning of global and local information. It is characterized by the integration of a human cognition-inspired module and a parallel haze constraint module, achieving robust dehazing performance under cloud and haze conditions. While it also adopts a multi-scale, global-local collaborative learning structure similar to ours, our method distinguishes itself by focusing on structure boundary restoration through the integration of SwinRRDB and a watershed-based structure-preserving loss function.

## 3. Proposed Method

### 3.1 Network Architecture

The following is a detailed explanation of the structure and proposed method of SUFERNOBWA. This study proposes a hybrid U-Net-based network to effectively remove atmospheric interference in satellite images. Conventional CNN-based U-Net models are efficient in learning local features, but they have limitations in handling large-scale atmospheric interference. Conversely, transformer-based models have been shown to effectively learn global contextual information, but they are susceptible to increased computational complexity. To address these issues, this paper proposes Swin Transformer-based Residual-in-Residual Dense Block (SwinRRDB) in the encoder( $E$ ) and decoder( $D$ ), while the bottleneck( $B$ ) is optimized with Bottleneck blocks, excluding Swin Transformer operations to balance local and global feature learning while maintaining computational efficiency. The overall structure follows the Residual-in-Residual Dense Block (RRDB) method, and internally, the Swin Transformer layer is applied three times to extract and refine features.

The SUFERNOBWA network is predicated on the U-Net architecture, whereby an RGB satellite image of dimensions 256×256×3 is accepted as input and an output image of identical dimensions is generated. The network is composed of three primary components: the encoder, the bottleneck, and the decoder (see Figure 1). The encoder progressively reduces the resolution while extracting key features, the bottleneck captures global contextual information, and the decoder restores the image to its original resolution using the learned features. The overall image dehazing process follows a pipeline, where the input image is first processed by the encoder to extract hierarchical features, then passed through the bottleneck to capture global contextual information, and finally reconstructed by the decoder. This process can be formally expressed (1), where $I_{\text{input}} \in \mathbb{R}^{256 \times 256 \times 3}$ represents the RGB satellite image input, and $\hat{I} \in \mathbb{R}^{256 \times 256 \times 3}$ denotes the final dehazed output image. This means that the proposed encoder, bottleneck, and decoder structures are sequentially applied to the input image to generate the output image.

$$\hat{I} = D\left(B\left(E(I_{\text{input}})\right)\right) \quad (1)$$

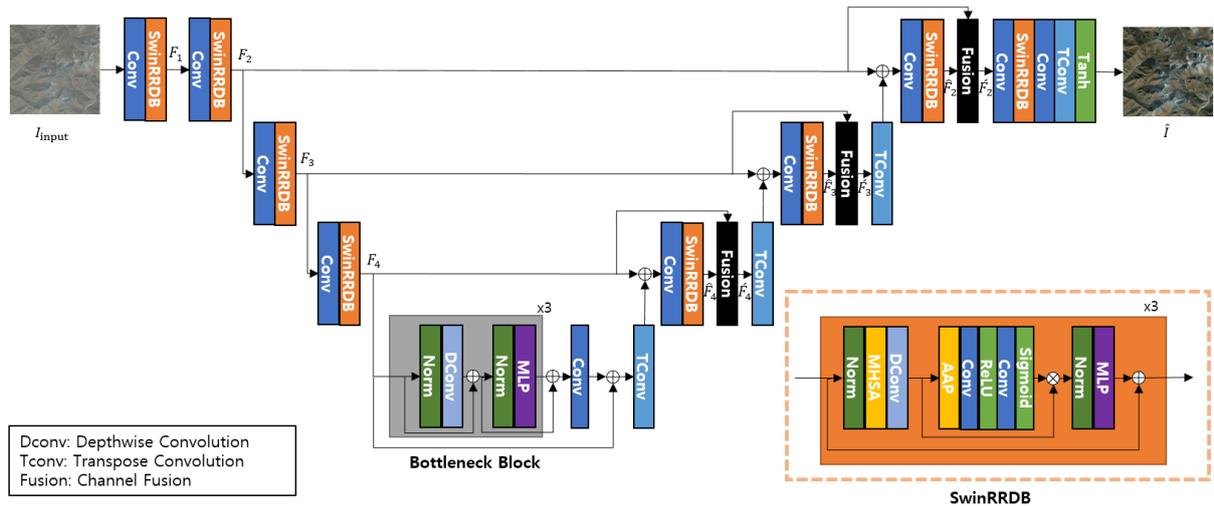

Fig. 1. Overall Architecture of the SUFERNOBWA Network for Image Dehazing

The encoder extracts meaningful features from the input image while progressively reducing the resolution to preserve essential information. The encoder is composed of four downsampling blocks, where each block halves the spatial resolution while increasing the number of feature channels to extract deeper features effectively. The encoding process follows these steps. First, a 3×3 convolution is applied to the input image to extract initial features, increasing the number of channels to facilitate deeper feature learning. After the initial convolution operation, three consecutive SwinRRDB are applied to simultaneously learn spatial and global features. The SwinRRDB are comprised of Multi-Head Self-Attention (MHSA), Depthwise Convolution, Attention, and an MLP, which collectively enable the effective capture of both spatial and global contextual information. Consequently, the final block in the encoder extracts the deepest-level features, which are subsequently fed into the bottleneck to facilitate further global feature learning. This process can be formally expressed (2). This means that the result of applying a 3x3 convolution to $F_{l-1}$ is fed into the SwinRRDB, and this process is performed across four layers.

$$F_l = \text{SwinRRDB}^{(l)}\big(\text{Conv}_{3\times 3}(F_{l-1})\big), \quad l \in \{1, 2, 3, 4\} \tag{2}$$

The bottleneck is the most critical stage of the SUFERNOBWA network, responsible for learning global contextual information in the deepest layer of the network. In conventional U-Net structures, the bottleneck is typically composed of elementary convolution layers, which serve to transform features. However, in order to minimize computational overhead and optimize global feature learning, this study employs Bottleneck blocks instead of Swin Transformer operations in the bottleneck. In summary, the bottleneck exclusively applies Bottleneck blocks without any additional convolutional operations.

The bottleneck block processes feature maps with the lowest spatial resolution in the network, and due to its relatively small input size, it is generally assumed that applying a self-attention mechanism in this stage would not significantly increase the overall computational complexity or memory usage. However, in this study, self-attention operations based on the Swin Transformer are deliberately omitted from the bottleneck. Instead, a lightweight bottleneck block composed solely of Depthwise Convolution and MLP modules is employed. This design choice further reduces the computational burden in the bottleneck while relying on the encoder and decoder to perform sufficient global feature learning via self-attention. In particular, the self-attention in the Swin Transformer is window-based and is primarily applied within the SwinRRDB blocks in both the encoder and decoder. This allows the network to maintain a large global receptive field while simplifying the bottleneck structure for improved efficiency and architectural balance.

This bottleneck block is primarily designed to efficiently refine and preserve the global and local features that have already been learned by the encoder and decoder. Specifically, Depthwise Convolution independently processes spatial information across each channel, making it effective for preserving local structural details. Following this, the MLP module facilitates high-dimensional feature refinement by modeling inter-channel interactions, thereby indirectly enhancing global contextual representations. In addition, the application of residual connections with a scaling factor ($\alpha = 0.1$) helps prevent overfitting and ensures stable feature propagation. This architectural configuration minimizes information loss within the bottleneck while suppressing computational overhead and maintaining the continuity of the global receptive field and consistency of feature flow throughout the network.

The decoder restores the image to its original resolution by integrating multi-resolution feature maps from the encoder and global contextual information from the bottleneck. The decoder is composed of four upsampling blocks, with each block doubling the resolution while progressively restoring image details. The decoding process begins with an upsampling operation to increase the resolution of the feature map. Before merging with the corresponding encoder feature map through a skip connection, a 1×1 convolution is applied to adjust the number of channels. The skip connection undergoes bilinear interpolation to match the upsampled feature's resolution. Thereafter, the two feature maps are concatenated along the channel dimension to form a unified feature representation. The merged feature map subsequently undergoes a 3×3 convolution, a process that eliminates unnecessary noise and refine the features. Subsequent to this refinement, SwinRRDB are applied to facilitate effective feature transformation. This process can be formally expressed (3).

$$\hat{F}_l = \text{SwinRRDB}^{(l)}\big(\text{Conv}_{3\times3}(\text{TConv}_{3\times3}(F_l) \oplus F_{l-1})\big), l \in \{4, 3, 2\} \quad (3)$$

Subsequently, the Channel Fusion process is performed. The purpose of this process is to effectively merge feature maps from different networks, thereby ensuring that the fusion process maintains spatial and semantic consistency. This is a crucial factor for improving image restoration and dehazing performance. First, two input feature maps, $F_l$ and $\hat{F}_l$, are concatenated along the channel dimension to create a joint representation. A 1×1 convolution is then applied to transform the concatenated features and extract meaningful feature interactions. This transformation helps reduce redundancy while capturing complementary information from both feature maps. Next, $A_{\text{weights}}$ applied to determine the relative importance of each feature map. It is a simple module that sequentially applies a convolution followed by a sigmoid function. In this process, $F_l$ represents the output of the SwinRRDB, which extracts deep and refined features by applying Swin Transformer-based residual dense learning. $\hat{F}_l$ corresponds to the output of the encoder at the same hierarchical level, containing essential structural and spatial information extracted from the input image. These two feature maps are integrated using an adaptive attention-based fusion mechanism. A second 1×1 convolution followed by a sigmoid activation function generates weights. Finally, the module generates the fused output $\acute{F}_l$ through the weighted sum of the two input feature maps. The output is formulated as (4).

$$\acute{F}_l = F_l \cdot A_{\text{weights}} + \hat{F}_l \cdot (1 - A_{\text{weights}}), l \in \{4, 3, 2\} \quad (4)$$

The final decoder layer employs a Conv-SwinRRDB-Conv-TransposeConv-Tanh structure to generate the final RGB output. This configuration facilitates the incorporation of Swin Transformer-based global contextual information, while the Tanh activation function ensures that the output pixel values fall within the [−1,1] range, thereby generating the final dehazed image.

3.1.4. SwinRRDB

In this study, SwinRRDB, which include the Swin Transformer, are utilized in both the encoder and decoder, while the bottleneck blocks exclude the Swin Transformer. The key processing blocks of this network are SwinRRDB and Bottleneck. The differences between these blocks and their processing steps are explained in detail below, including relevant equations. SwinRRDB is used in both the encoder and decoder and is designed to learn global contextual information by incorporating the Swin Transformer. SwinRRDB extends the architecture of the Swin Transformer by incorporating the Residual-in-Residual Dense Block (RRDB) scheme. Unlike simply applying Self-Attention, SwinRRDB further enhances feature representations through multiple layers of residual connections and dense connections. $F_{\text{RRDB}}$ represents the input feature map, and $F_{\text{conv}}$ is the output feature map after passing through the Swin Transformer layers and subsequent operations. To enhance training stability and alleviate vanishing gradient issues in deep networks, the output feature map $F_{\text{conv}}$ is scaled by a residual factor α, which is set to 0.2. In SwinRRDB, Layer Normalization is first applied to the input feature map $X$, followed by Multi-Head Self-Attention (MHSA) in the Swin Transformer. Given the input $X$, the MHSA operation is defined as (5). $W_Q, W_K, W_V$ are weight matrices that generate queries($Q$), keys($K$), and values($B$), and $d_k$ is a scaling factor. The attention scores allow learning of global relationships between input data, which is then locally applied using the window-based mechanism of the Swin Transformer.

$$Q = W_Q X, K = W_K X, \quad V = W_V X,$$
$$F_{attn} = \text{Attention}(Q, K, V) = \text{softmax}\left(\frac{QK^T}{\sqrt{d_k}}\right)V \quad (5)$$

To further refine features extracted by the Swin Transformer, Depthwise Convolution is applied. Channel attention is a process that adjusts the importance of each channel, and is computed using an algorithm called Adaptive Average Pooling (AAP). This is followed by two consecutive 1×1 convolutions.

It is defined as (6). $W_1, W_2$ are weights of the 1×1 convolutions, and σ is the sigmoid activation function. And the result of applying depthwise convolution is denoted as $F_{\text{dwc}}$. Finally, an MLP module $F_{\text{mlp}}$ is applied for feature transformation, expressed as (7). The GELU (Gaussian Error Linear Unit) function is a nonlinear activation function that operates by probabilistically and selectively allowing input values to pass through. The larger the input value, the higher the probability of it being passed, while smaller values are more likely to be suppressed. It offers a smoother and more continuous curve compared to the ReLU function. Since it adjusts the input smoothly based on the cumulative distribution function (CDF) of the normal distribution, it contributes to faster convergence during training and enhances the representational capacity of the model.

$$F_{\text{c}} = \sigma\left(W_2 * \text{ReLU}(W_1 * \text{AAP}(F_{\text{dwc}}))\right) \quad (6)$$

$$F_{\text{mlp}} = W_2 \cdot \text{GELU}(W_1 * F_{\text{c}}),$$
$$\text{GELU}(W_1 * F_{\text{c}}) = W_1 * F_{\text{c}} * CDF(W_1 * F_{\text{c}}) \quad (7)$$

The SwinRRDB block is designed to jointly learn global contextual and local spatial features by integrating the window-based Multi-Head Self-Attention (MHSA) mechanism from the Swin Transformer with the Residual-in-Residual Dense Block (RRDB) architecture from CNNs. Initially, the input feature map is partitioned into fixed-size windows, and self-attention is computed within each window to capture local contextual relationships. The use of window shifting across layers enables interaction between adjacent windows, thereby expanding the global receptive field across the entire feature space. Following self-attention, Depthwise Convolution is applied to compensate for the potential loss of fine spatial details that attention mechanisms might overlook. By independently convolving each channel, this operation effectively preserves structural features, such as building boundaries and road edges, which are critical in remote sensing imagery. Furthermore, Channel Attention and MLP layers are employed to model inter-channel dependencies and emphasize semantically important features. The combination of these components allows SwinRRDB to simultaneously integrate global semantic context and precise spatial structures.

### 3.2 Novel Loss Functions

In the proposed image dehazing framework, a comprehensive loss function is designed to ensure that the restored image maintains both pixel-wise accuracy and structural integrity. Conventional loss functions frequently encounter limitations in fully recovering fine details and sharp edges, which are paramount for generating high-quality dehazed images. To address this limitation, the proposed loss function integrates three distinct components: L2 Loss, which minimizes pixel-wise reconstruction error, guided loss, which enhances texture consistency, and watershed loss, which preserves structural boundaries. Each of these loss terms plays a crucial role in improving the quality of the dehazed image by focusing on different aspects of the restoration process.

The first component, L2 Loss $\mathcal{L}_{\text{L2}}$, serves as a fundamental constraint in the learning process. It is defined as the squared difference between the predicted dehazed image and the ground truth image. By penalizing large deviations more heavily than small ones, L2 Loss encourages the model to produce outputs that are numerically closer to the ground truth. The formulation of L2 Loss is given as (8). Where $I_{\text{pred}}^{(i)}$ represents the intensity of pixel $i$ in the predicted dehazed image, and $I_{\text{GT}}^{(i)}$ is the corresponding pixel intensity in the ground truth image. The loss is averaged over all pixels $N$ in the image. Although L2 Loss ensures numerical similarity to the reference image, it does not explicitly preserve fine textures or edge details, often resulting in over-smoothed outputs.

$$\mathcal{L}_{\text{L2}} = \frac{1}{N}\sum_{i=1}^{N}\left(I_{\text{pred}}^{(i)} - I_{\text{GT}}^{(i)}\right)^2 \quad (8)$$

To mitigate texture loss due to over-smoothing, guided loss is introduced as an additional constraint. The guided filter [43] is an edge-preserving filter that enforces structural consistency between an image and a guiding reference. In the context of dehazing, the guided filter is applied to both the ground truth and the predicted image, ensuring that fine textures remain intact while maintaining smooth transitions. The filtering process involves computing local means, variances, and covariances to derive a linear transformation that aligns the restored image with the reference structure. $\mu_{GT}$ and $\mu_{pred}$ are applied mean filtering, respectively, and $a$ and $b$ are the linear coefficients computed for guided filtering. The final guided filter output $I_{guided}$ is computed as (9). In the given expression, $t \in \{GT, pred\}$ indicates that $t$ and $\bar{t}$ can either represent the ground truth (GT) or the predicted value (pred). $I_{guided,GT}$ refers to the guided image derived from the GT and $I_{guided,pred}$ refers to the guided image derived from the pred.

$$I_{guided,t} = a(t) \cdot I_t + b(t),$$

where $a(t) = \frac{I_t \cdot I_{GT} - \mu_t \cdot \mu_{GT}}{\sigma_t^2 + \epsilon}, b(t) = \mu_{GT} - a(t) \cdot \mu_t,$

$$t \in \{GT, pred\} \qquad (9)$$

Once the guided-filtered versions of the ground truth and predicted images are obtained, the guided loss $\mathcal{L}_{guided}$ is defined as (10). where $I_{guided,GT}$ and $I_{guided,pred}$ are the guided-filtered versions of the ground truth and predicted image, respectively. By incorporating this loss, the network is encouraged to retain fine textures that would otherwise be lost due to over-smoothing.

$$\mathcal{L}_{guided} = \frac{1}{N}\sum_{i=1}^{N}\left|I_{guided,GT}^{(i)} - I_{guided,pred}^{(i)}\right| \qquad (10)$$

While L2 Loss and guided Loss focus on pixel similarity and texture consistency, they do not explicitly enforce the preservation of object boundaries, which is crucial for high-quality image restoration. To address this, watershed loss is introduced as the third component of the loss function. The watershed algorithm is a classical image segmentation method that treats pixel intensity variations as topographic surfaces, where local minima serve as basin regions. The application of this algorithm to both the ground truth and the predicted dehazed image enables the extraction of structural boundaries, which can then be utilized as a loss constraint. The watershed transformation follows a multi-step process.

Satellite imagery typically exhibits a high density of edges due to the presence of diverse structures such as roads and buildings. In such environments, directly applying a conventional watershed algorithm leads to excessive computational complexity and high sensitivity to parameter tuning, making its practical application challenging. Edge extraction methods, such as high-frequency enhancement, offer a relatively low computational cost and enable rapid identification of major boundaries. However, extracting only edge information is insufficient for achieving complete object-level segmentation. In contrast, region-based segmentation using the watershed algorithm requires significantly higher computational resources compared to simple edge detection. Specifically, processes such as local minima detection, marker generation, and iterative region expansion substantially increase the overall computational burden. To address these issues, this study adopts an alternative approach by applying Gaussian smoothing instead of enhancing high-frequency components. By removing unnecessary fine edges while preserving major structural boundaries, the preprocessed image enables the watershed algorithm to perform stable and meaningful region segmentation. Consequently, this strategy achieves a substantial reduction in computational cost while maintaining the structural integrity essential.

The proposed process for generating the watershed label map involves the following key steps:

**Gaussian Smoothing**: The watershed algorithm is a representative image segmentation method that treats pixel intensities as topographic elevations, segmenting regions based on local minima. While it typically operates on a gradient map, satellite imagery and structurally complex images often contain an excessive number of fine edges, resulting in over-segmentation and high sensitivity to noise. To solve these issues, Gaussian smoothing is applied before segmentation. This process attenuates high-frequency components, suppressing irrelevant fine edges while preserving essential structural boundaries, thereby enabling more stable and robust watershed segmentation.

**Local Minima Detection and Marker Assignment**: From the smoothed image, local intensity minima are detected using a neighborhood-based search. Each detected minimum is assigned a unique integer label, which serves as a seed marker for watershed propagation. This ensures that segmentation begins from structurally meaningful and spatially stable regions, laying a reliable foundation for subsequent label expansion.

**Four-Directional Label Propagation**: Based on the initialized markers, a region growing process is performed using four-directional label propagation (up, down, left, and right). At each iteration, neighboring unlabeled pixels with minimal intensity difference from the source label region are assigned the same label. This iterative expansion continues until all pixels are labeled, allowing for smooth region boundaries while effectively suppressing propagation noise.

**Label Map Normalization**: Since the number and range of labels may vary across images, normalization is essential to ensure consistent loss computation. The raw label map is normalized to the range [0, 1] using the formulation defined in Equation (11). $I_{\text{water}}$ represents the raw watershed label map, and $\epsilon$ prevents division by zero. The watershed loss is then computed as the L2 Loss between the normalized label maps of the ground truth and predicted image. It is defined as (12).

$$I_{\text{water}} = \frac{I_{\text{water}} - \min(I_{\text{water}})}{\max(I_{\text{water}}) - \min(I_{\text{water}}) + \epsilon} \quad (11)$$

$$\mathcal{L}_{\text{water}} = \frac{1}{N} \sum_{i=1}^{N} \left( I_{\text{water,pred}}^{(i)} - I_{\text{water,GT}}^{(i)} \right)^2 \quad (12)$$

When using only boundary information, it is possible to capture the outlines of objects within an image; however, there are inherent limitations in fully restoring the overall structure. Particularly in satellite imagery, where multiple complex structures coexist, boundary information alone is insufficient to clearly distinguish between different objects or to accurately reconstruct their shapes. Therefore, in this study, we adopt an approach that utilizes not only boundary information but also label information that defines each object as an independent region. The label map generated through the Watershed algorithm provides not merely outline information but also ensures inter-object separation and internal structural consistency. This enables clearer preservation of boundaries between key structures such as roads, buildings, and rivers during the restoration process, and prevents distortion of object shapes. Thus, incorporating label information is essential for improving structural restoration accuracy and enhancing the practical utility of the restored images. $I_{\text{water,GT}}$ and $I_{\text{water,pred}}$ are the normalized watershed label maps of the ground truth and predicted image, respectively.

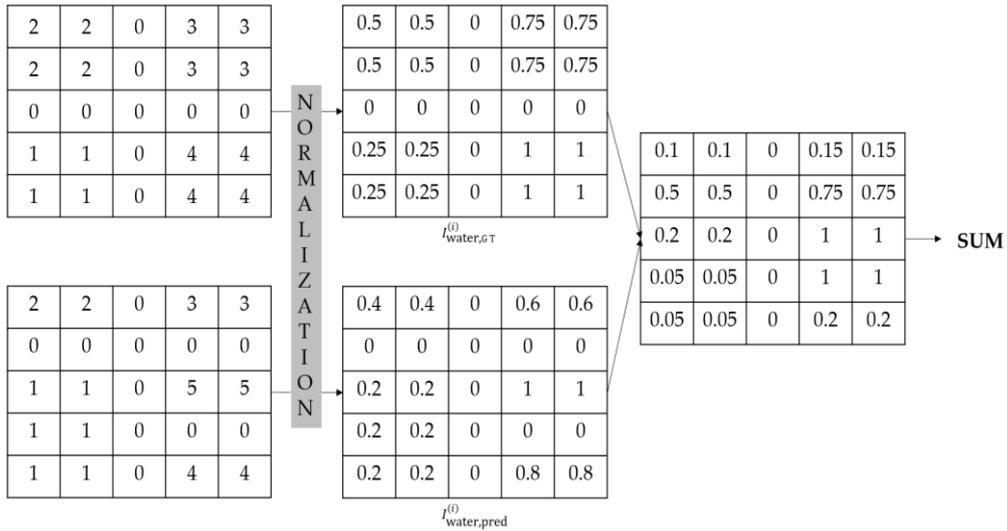

Fig. 2. Computation Process of the Proposed Loss Function Using the Watershed Algorithm

Figure 2 visually illustrates the process of calculating the proposed loss function, $\mathcal{L}_{\text{water}}$. First, the Watershed label maps mentioned above are generated for both the input image and the output image. Each of these label maps is then normalized based on its maximum value. The normalized maps are used to compute the pixel-wise L1 difference, resulting in a difference map. The final loss is defined as the sum of the remaining label values in this difference map. If the difference between the input and output images is small, this sum value will also be small, indicating higher structural similarity.

Finally, the total loss function $\mathcal{L}_{\text{total}}$ is defined as a weighted sum of these three components in (13). $\lambda_1$, $\lambda_2$, and $\lambda_3$ are weighting coefficients that balance the contribution of each loss term.

$$\mathcal{L}_{\text{total}} = \lambda_{\text{L2}}\mathcal{L}_{\text{L2}} + \lambda_{\text{guided}}\mathcal{L}_{\text{guided}} + \lambda_{\text{water}}\mathcal{L}_{\text{water}} \tag{13}$$

By minimizing this total loss function, the proposed dehazing model achieves a balance between pixel fidelity, texture preservation, and structural integrity. L2 Loss ensures accurate reconstruction, guided loss enforces texture consistency, and watershed loss maintains object boundaries, collectively leading to high-quality image dehazing.

## 4. Experiments and Results Analysis

### 4.1 Datasets

In this paper, two datasets, RICE [44] and SateHaze1k [22], are used to evaluate the proposed method. RICE is a remote sensing image dataset designed for cloud removal, consisting of 500 paired images, where each pair consists of a cloudy image and a cloud-free image with a resolution of 512×512 pixels. In this study, 390 pairs are used for training and 110 pairs are used for testing. The SateHaze1k dataset contains hazy images, their corresponding clear images, and SAR images provided in paired format with a resolution of 512×512 pixels. This dataset contains three levels of haze: thin, moderate, and thick. The training set consists of 900 mixed pairs covering different haze levels, while the test set contains 45 pairs for each haze level. For consistency and computational efficiency, all datasets are resized to 256×256 pixels for processing in this study.

### 4.2 Environments and Training Information

In this study, the experimental environment was based on the Ubuntu 18.04 LTS operating system, and two NVIDIA RTX 3090 GPUs were used in parallel to efficiently perform large-scale computations. his configuration optimized training and evaluation speed, allowing for more efficient model training. During training, the Adam Optimizer was employed, with an initial learning rate set to 1e-5 to ensure stable learning. The training process was conducted for a total of 1000 epochs, allowing the model to gradually optimize through sufficient iterative learning. The hyperparameters were set as follows: $\lambda_{\text{L2}} = 5$, $\lambda_{\text{guided}} = 1$, and $\lambda_{\text{water}} = 0.5$. These parameters were adjusted to enhance the model's generalization performance and adapt to specific data characteristics. $\lambda_{\text{L2}}$ helps prevent overfitting and promotes better generalization, while $\lambda_{\text{guided}}$ and $\lambda_{\text{water}}$ facilitate more effective learning by reflecting specific patterns in the data.

The performance of the model was evaluated using Peak Signal to Noise Ratio (PSNR) and Structural Similarity Index Measure (SSIM). PSNR quantifies the ratio of signal to noise and provides a metric for assessing the difference between the restored and original images. A higher PSNR value indicates a higher quality restoration, implying greater similarity to the original image. SSIM measures the structural similarity between two images, considering human visual perception. A value closer to 1 indicates that the restored image retains the structural integrity of the original, implying more reliable reconstruction performance.

### 4.3 Result Analysis

A quantitative evaluation was performed on the RICE dataset, and as shown in Table 1, the method proposed in this study (Ours) achieved a PSNR of 33.24 dB and an SSIM of 0.967, surpassing all the approaches compared. First, RSDformer [24] achieved a relatively high PSNR of 33.01 dB and an SSIM of 0.953, but compared to our method, the PSNR differs by about 0.23 dB and the SSIM by 0.014. In the field of satellite image restoration, even a 0.1 dB improvement in PSNR can be considered visually significant, indicating that our method clearly exceeds that threshold.

In addition, J. Hwang et al. [26] show a PSNR of 28.90 dB and an SSIM of 0.910, revealing a large gap in terms of both pixel-level accuracy and structural preservation compared to our approach, while DehazeNet [28] also posts relatively low values (PSNR 29.48 dB, SSIM 0.921). This finding suggests that CycleGAN-based or traditional end-to-end networks do not sufficiently capture the structural information needed to handle the complex atmospheric or cloud interferences commonly present in satellite and aerial imagery.

Meanwhile, S. Vishwakarma et al. [30] exhibits a PSNR of 27.34 dB and an SSIM of 0.912, the lowest PSNR among the compared methods, which implies that, despite being a relatively recent study, it fails to adequately preserve boundary and high-frequency details when restoring real satellite images. HALP [31] demonstrates a PSNR of 20.91 dB and an SSIM of 0.926; although the SSIM is at a moderate level, the large pixel-level error leads to a significantly reduced PSNR. SCANet [32] achieves a PSNR of 30.04 dB and an SSIM of 0.893, but lags behind the proposed method in maintaining boundaries and structures, resulting in comparatively lower PSNR and SSIM scores. SpA GAN [34] achieved a PSNR of 30.23 dB and an SSIM of 0.954, demonstrating the importance of accurately preserving boundary information for terrain and buildings, much like the proposed method. DehazeFormer [35] achieves a PSNR of 32.55 dB and an SSIM of 0.931, demonstrating solid performance in high-resolution dehazing. However, compared to the proposed method, it is somewhat inferior in preserving boundary and structural details.

C. Li et al. [38] achieved a PSNR of 27.08 dB and an SSIM of 0.940. Their method incorporates Gaussian-weighted image fusion and unsharp mask-based color correction to enhance transmittance estimation accuracy while addressing color distortion issues. However, both the PSNR and SSIM values are lower compared to our method (PSNR 33.24, SSIM 0.967), indicating the limitations of traditional restoration and enhancement-based approaches in handling the complex structures and color information present in satellite imagery. UAVD-Net [39] demonstrated a PSNR of 32.05 dB and an SSIM of 0.931. While UAVD-Net shows competitive performance in structural restoration and color enhancement through its global-local feature integration, its SSIM of 0.931 is lower than that of our method, and its PSNR is 1.19 dB lower than our method's 33.24 dB. This demonstrates that although UAVD-Net effectively captures global and local features, our integration of SwinRRDB with a watershed-based structure-preserving loss allows for more effective restoration of fine structural details and consistency in color, resulting in superior reconstruction accuracy.

Table 1. Performance Comparison Using the RICE Dataset (PSNR, SSIM).

|  | PSNR | SSIM |
|---|---|---|
| RSDformer [24] | 33.01 | 0.953 |
| J. Hwang et al. [26] | 28.90 | 0.910 |
| DehazeNet [28] | 29.48 | 0.921 |
| S. Vishwakarma et al. [30] | 27.34 | 0.912 |
| HALP [31] | 20.91 | 0.926 |
| SCANet [32] | 30.04 | 0.893 |
| SpA GAN [34] | 30.23 | 0.954 |
| DehazeFormer [35] | 32.55 | 0.931 |
| C. Li et al. [38] | 27.08 | 0.940 |
| UAVD-Net [39] | 32.05 | 0.931 |
| K. He et al. [45] | 16.03 | 0.697 |
| Ours | **33.24** | **0.967** |

Ultimately, in challenging image restoration tasks such as atmospheric or cloud removal, merely minimizing pixel-level differences is insufficient. Preserving edges and structural features plays a pivotal role. By effectively incorporating structural and boundary information while also maximizing pixel-level restoration accuracy through its network and loss function design, our proposed method achieves superior PSNR and SSIM. In particular, it demonstrates the ability to simultaneously control structural distortions and pixel errors in high-resolution datasets with complex detail, such as satellite and aerial imagery. This achievement significantly increases its practical value for various remote sensing applications, including disaster monitoring, urban and agricultural planning, and environmental surveillance. To further investigate the limitations of traditional physics-based dehazing methods, a quantitative analysis was conducted using the representative approach proposed by K. He et al. [45]. On the RICE dataset, this method achieved a PSNR of 16.03 and an SSIM of 0.697. These results indicate that even under relatively uniform and simple haze conditions, the restoration performance is suboptimal, with insufficient preservation of structural consistency. In the context of high-resolution satellite imagery, where texture and boundary information play a critical role in downstream analysis, such performance levels pose significant limitations for practical deployment.

Figure 3 presents a visual quality comparison between the input images, SCANet [32], SpAGAN [34], SUFERNOBWA, and Ground Truth (GT). The analysis results indicate that the proposed method achieves the best performance in terms of color preservation, structural restoration, and cloud/fog removal. First, in terms of color and brightness preservation, SCANet [32] and SpAGAN [34] exhibited some degree of color distortion after cloud removal. Specifically, in vegetated areas (Rows 1 and 2), some methods applied excessive color correction, resulting in an unnaturally emphasized green hue. In mountainous regions (Row 3), certain methods produced overly bright or yellowish tones. In contrast, the proposed method maintains the most similar color tone to the Ground Truth while naturally restoring brightness and contrast, demonstrating its superiority. Next, in structural and boundary restoration, SCANet [32] and SpA GAN [34] tended to blur the boundaries of roads and rivers. However, the proposed method preserves boundaries naturally while maintaining the original structure of the images, delivering the most stable structural restoration performance. Particularly in mountainous regions (Row 3), the proposed method accurately restores terrain contours and boundaries, outperforming other approaches. Finally, in cloud and fog removal performance, SCANet [32] and SpAGAN [34] exhibited residual haze, limiting their ability to fully restore the image. In contrast, the proposed method effectively removes clouds and fog while preserving both color and structural information, generating results most similar to the Ground Truth. In conclusion, the proposed method demonstrates the best performance in color preservation, structure retention, and boundary restoration, overcoming the limitations of existing approaches. In particular, its superior performance in restoring colors and boundaries in mountainous areas, as well as providing natural results after cloud and fog removal, highlights its high potential for real-world applications.

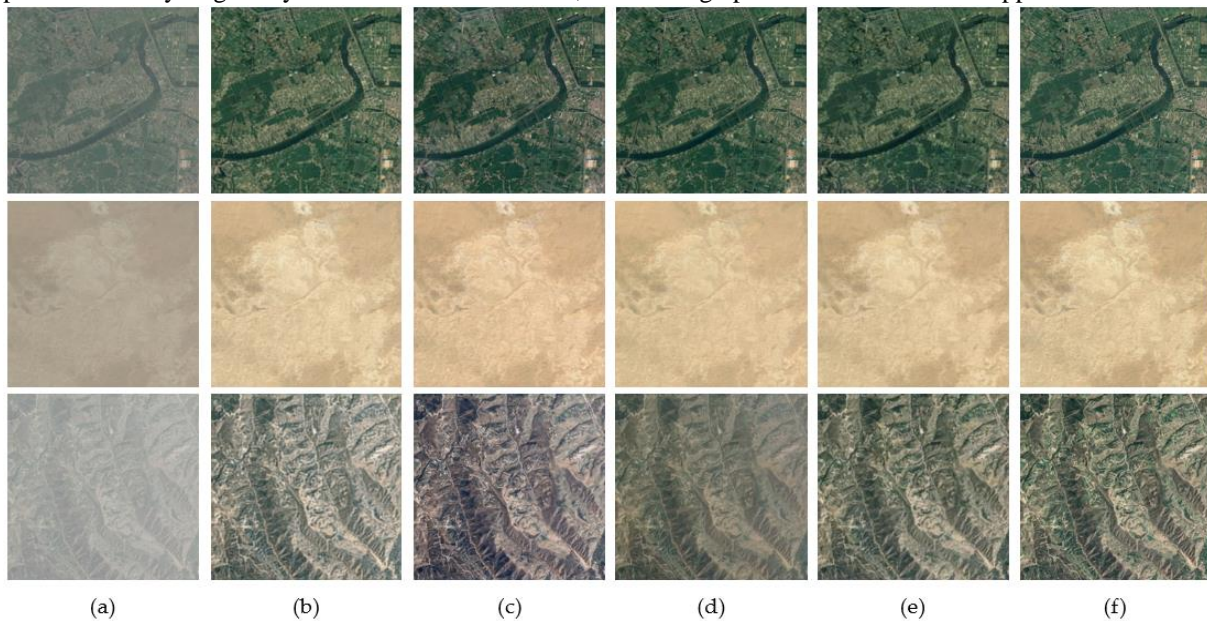

Fig. 3. Comparison of Generated Results Using the RICE Dataset ((a): Input, (b): SCANet [32], (c): SpAGAN [34], (d): DehazeFormer[35], (e): SUFERNOBWA, (f): Ground Truth)

Figure 4 presents a comparative evaluation of the proposed method's performance by comparing the input image, restored image, and Ground Truth using RICE datasets [34]. The figure consists of four rows, representing urban area with a river, urban area in a desert, desert region, and vegetated region. The analysis indicates that the proposed method exhibits superior cloud and haze removal performance, excelling in both structural preservation and color fidelity. In the urban area with a river (Row 1), the river boundaries are clearly restored, and the surrounding urban structures are accurately reconstructed. In the urban area in a desert (Row 2), building layouts and road patterns are distinctly restored, while brightness and contrast are naturally maintained. In the desert region (Row 3), terrain contours are sharply recovered, and the method successfully preserves the original color balance without excessive color correction. Lastly, in the vegetated region (Row 4), vegetation and man-made structures are restored with high consistency relative to the Ground Truth. Thus, the proposed method demonstrates robust cloud removal and structural restoration capabilities across various terrain environments, proving to be an effective and optimized solution for remote sensing image restoration.

Subsequently, we evaluated the performance of the proposed method and existing techniques using the Satellite Haze1k [22] dataset, which classifies atmospheric conditions into Thin Fog, Moderate Fog, and Thick Fog. The quantitative results are summarized in Table 2. Under Thin Fog conditions, X. Chen et al. [23] achieved PSNR 25.84 dB and SSIM 0.930, whereas the proposed method (Ours) showed a slightly lower PSNR of 24.19 dB but registered the highest SSIM at 0.949. This indicates that even in environments with minimal fog, our approach not only aligns pixel values but also preserves edge and structural information, ultimately yielding outcomes more advantageous for visual/geographical interpretation and subsequent data processing.

In the Moderate Fog regime, the superiority of the proposed method becomes more apparent. While other studies report PSNR values in the 25–26 dB range, our method achieves PSNR 28.15 dB and SSIM 0.950, securing the highest scores for PSNR metrics. For instance, DehFormer [25] reached PSNR 26.80 dB and SSIM 0.944, which, although competitive among previous works, is still 1.35 dB and 0.006 lower than the proposed method. This strongly demonstrates the efficacy of our innovative network architecture and loss function design, particularly when the scene contains a mix of low-frequency and high-frequency components under moderate fog conditions.

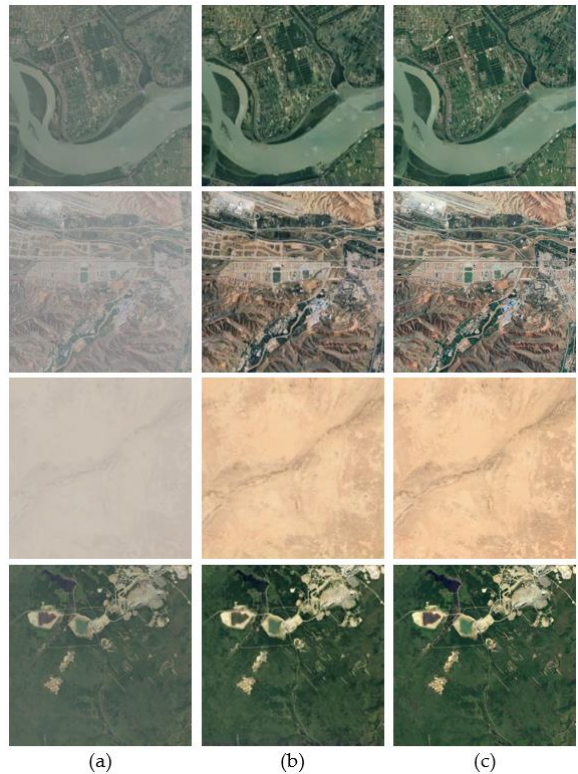

Fig. 4. Generated Results of SUFERNOBWA Using the RICE Dataset ((a): Input, (b): SUFERNOBWA, (c): Ground Truth)

Even in the Thick Fog scenario, our method maintains the top structural similarity (SSIM = 0.910), despite some prior methods attaining marginally higher PSNR values. This phenomenon reflects the tendency of low-frequency-based restoration approaches to inflate PSNR in extremely dense fog at the expense of blurred boundaries or object merging (Over-Smoothing). Conversely, our method emphasizes high-frequency details and boundary preservation, thereby keeping object contours and overall structure crisp under severe fog.

Table 2. Performance Comparison Using the SateHaze1k Dataset (PSNR, SSIM).

|  | Thin Fog | | Moderate Fog | | Thick Fog | |
| --- | --- | --- | --- | --- | --- | --- |
|  | PSNR | SSIM | PSNR | SSIM | PSNR | SSIM |
| B. Huang et al. [22] | 24.16 | 0.906 | 25.31 | 0.926 | <u>25.07</u> | 0.864 |
| X. Chen et al. [23] | <u>25.84</u> | 0.930 | 26.15 | 0.928 | **25.20** | <u>0.901</u> |
| RSDformer [24] | 24.21 | 0.912 | 26.24 | 0.934 | 23.01 | 0.853 |
| DehFormer [25] | 24.49 | 0.919 | 26.80 | 0.944 | 22.24 | 0.854 |
| M2SCN [29] | 25.21 | 0.918 | 26.11 | 0.942 | 21.33 | 0.829 |
| HALP [31] | 19.79 | 0.899 | 18.86 | 0.909 | 16.17 | 0.808 |
| DehazeFormer [35] | 24.41 | 0.848 | 26.24 | 0.919 | 21.84 | 0.743 |
| PhDNet [36] | 23.13 | 0.897 | 26.47 | 0.942 | 22.95 | 0.882 |
| DVKT [37] | 24.73 | 0.918 | 27.05 | 0.942 | 23.31 | 0.886 |
| UDAVM-Net [40] | **26.76** | <u>0.934</u> | <u>27.53</u> | **0.952** | 23.48 | 0.873 |
| ICL-Net [41] | 24.59 | 0.923 | 25.67 | 0.937 | 21.78 | 0.859 |
| K. He et al. [45] | 13.15 | 0.725 | 9.78 | 0.574 | 10.25 | 0.585 |
| Ours | 24.19 | **0.949** | **28.15** | <u>0.950</u> | 22.33 | **0.910** |

UDAVM-Net [40] achieved PSNR values of 26.76, 27.53, and 23.48 dB, and SSIM values of 0.934, 0.952, and 0.873 under the thin, moderate, and thick haze conditions, respectively. Notably, under the moderate haze condition, it showed strong performance with a PSNR of 27.53 dB and an SSIM of 0.952. However, our proposed method achieved a PSNR of 28.15 dB and an SSIM of 0.950, indicating a 0.62 dB improvement in PSNR while the SSIM was slightly lower by 0.002, which is a negligible difference. ICL-Net [41] achieved PSNR values of 24.59, 25.67, and 21.78 dB, and SSIM values of 0.923, 0.937, and 0.859 under the thin, moderate, and thick haze conditions, respectively. Our proposed method outperformed ICL-Net with an SSIM of 0.949 under the thin haze condition and a PSNR of 28.15 dB under the moderate haze condition.

In the case of K. He et al. [45], the model achieved a PSNR of 13.15 dB and an SSIM of 0.725 under the Thin condition, but a significant performance drop was observed under the Moderate condition, with a PSNR of 9.78 dB and SSIM of 0.574, and under the Thick condition, with a PSNR of 10.25 dB and SSIM of 0.585. This suggests that as the density or distribution of haze changes, physics-based models fail to capture the increased complexity and are unable to deliver consistent restoration results.

In conclusion, the proposed method demonstrates balanced restoration performance across all fog densities—from thin to moderate and thick. In particular, it achieves the highest PSNR and SSIM in the Moderate Fog region, and ranks first in SSIM under both Thin Fog and Thick Fog conditions, showcasing outstanding capabilities for edge and structural preservation. Although PSNR may be somewhat lower in certain fog conditions, overall SSIM remains the highest, ensuring superior visual fidelity and the retention of fine-grained details in the restored images. These strengths imply that the proposed method can deliver stable and precise restoration performance in actual remote sensing operations and serve as the most effective, scalable solution across a diverse array of application domains.

Figure 5 compares the restoration results under the most challenging Thick (dense haze) condition of the SateHaze1K dataset. The experiment involves DehazeFormer [35], RSDFormer [24], and the proposed method, and their results are analyzed by comparing them with the input image and the ground truth (GT). DehazeFormer [35] and RSDFormer [24] exhibit overall color harmony and maintain a natural color tone, particularly showing regional color representation similar to the ground truth. In contrast, the proposed method tends to produce relatively brighter colors. However, in terms of structural restoration, while DehazeFormer [35] and RSDFormer [24] excel in color consistency, they show some deficiencies in restoring structural elements. Specifically, roads, buildings, and agricultural areas tend to appear blurry, and their boundaries may become distorted. On the other hand, the proposed method demonstrates outstanding structural similarity, effectively preserving clear boundaries and the shapes of buildings. Although the proposed method tends to generate brighter colors, its high structural similarity makes it more suitable for practical applications. In satellite image-based analysis, accurately restoring key features such as buildings and roads is crucial, and preserving structural information can be interpreted as a superior performance indicator. In conclusion, while DehazeFormer [35] and RSDFormer [24] achieve natural color reproduction with high color similarity to the ground truth, the proposed method outperforms them in structural restoration, preserving the shapes of key objects such as buildings and roads more clearly. Therefore, when considering the practical use of satellite images, where accurate terrain and structural recognition are essential, the proposed method is more suitable.

Figure 6 presents a comparative evaluation of restoration results using the SateHaze1k [22], where each column corresponds to hazy images, Restored images, and Ground Truth (GT). The rows represent different haze conditions: Row 1: Thin, Row 2: Moderate, Row 3: Thick. The Satellite haze1k dataset poses a higher level of restoration difficulty compared to RICE [44] due to its proximity to the ground and the non-uniform distribution of haze, making it more challenging to maintain both color fidelity and structural integrity during the restoration process. The analysis shows that the proposed method effectively preserves key structural elements such as buildings and roads while achieving stable restoration results. Particularly, in the Thick (Row 3) environment, the boundaries of buildings and roads remain relatively well-preserved, exhibiting less structural distortion compared to existing methods. However, in certain regions, the reflectance of rooftops appears lower than the original, and in the Thick environment, the overall color discrepancy is more pronounced.

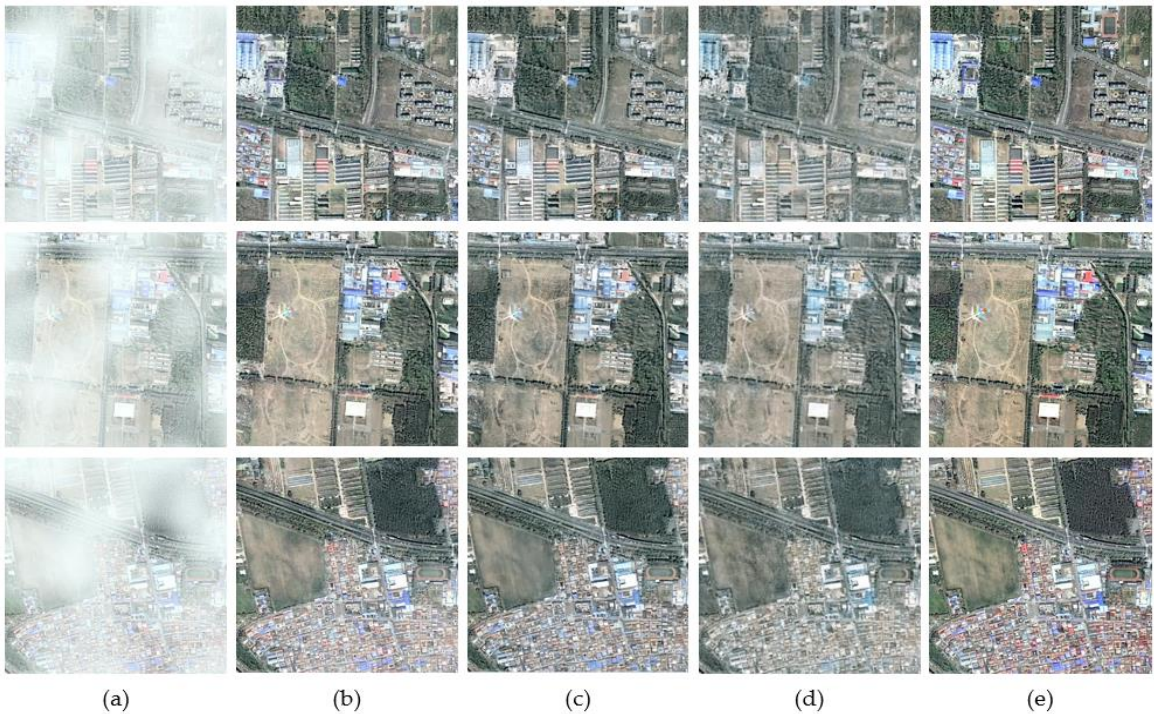

Fig. 5. Comparison of Generated Results Using Thick Fog of the SateHaze1K Dataset ((a): Input, (b): DehazeFormer [35], (c): RSDFormer [24], (d): SUFERNOBWA, (e): Ground Truth)

Additionally, road and edge details appear relatively less distinct compared to RICE [44] restoration results. This can be attributed to color reconstruction differences and the inherently limited information present in the input hazy images, as further reflected in the PSNR results. Nevertheless, when analyzed alongside quantitative evaluations, the proposed method demonstrates superior restoration performance, minimizing color distortion while preserving structural information with high stability. Consequently, it serves as an effective and robust solution for diverse haze conditions, proving its applicability in real-world remote sensing image restoration

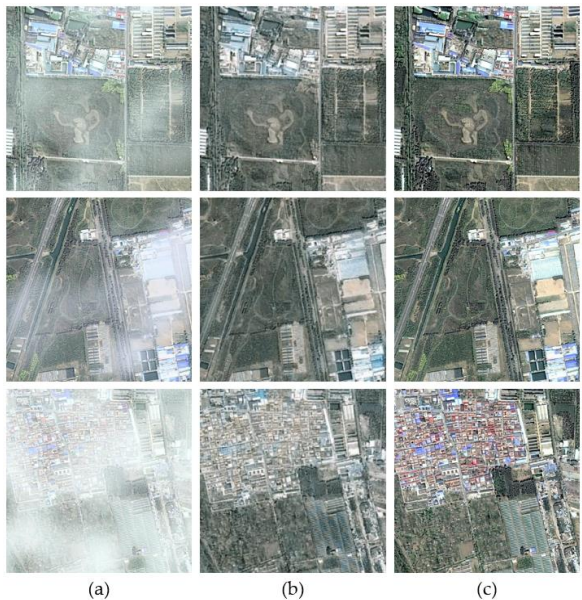

(a)  (b)  (c)

Fig. 6. Generated Results of SUFERNOBWA Using the SateHaze1k Dataset ((a): Input, (b): SUFERNOBWA, (c): Ground Truth)

### 4.4 Ablation Study

In this study, an Ablation Study was conducted to verify the effectiveness of the proposed loss functions using the RICE dataset [44]. To analyze the impact of each loss function, we evaluated performance variations by selectively applying Guided Loss and Watershed loss, while keeping the L2 Loss fixed as a baseline. The dehazing performance was assessed using PSNR (Peak Signal-to-Noise Ratio), SSIM (Structural Similarity Index Measure), and UQI (Universal Quality Index). UQI integrates luminance, contrast, and structural information into a single scalar metric, and evaluates overall image quality based on the interrelationship among these three components. The experimental results are presented in Table 3.

First, when using only L2 Loss, the model trained on the RICE dataset achieved PSNR of 32.71 dB, SSIM of 0.947, and UQI of 0.791. This result indicates that when the model is trained solely to minimize pixel-wise differences, its dehazing performance remains limited. L2 Loss focuses on minimizing per-pixel errors, but in image restoration tasks such as dehazing, it struggles to preserve structural information and edge details effectively. The proposed model architecture demonstrates high PSNR performance even when utilizing only L2 Loss. However, it was observed that some outputs exhibit blurring or over-smoothing artifacts.

Table 3. Performance Comparison Based on the Usage of Loss Functions (PSNR, SSIM, UQI).

| $\mathcal{L}_{L2}$ | $\mathcal{L}_{guide}$ | $\mathcal{L}_{water}$ | PSNR | SSIM | UQI |
|---|---|---|---|---|---|
| O | X | X | 32.71 | 0.947 | 0.791 |
| O | X | O | 32.28 | 0.965 | 0.833 |
| O | O | X | 32.61 | 0.966 | 0.827 |
| O | O | O | **33.24** | **0.967** | **0.835** |

Adding Watershed Loss significantly improved performance, increasing PSNR to 32.28 dB, SSIM to 0.965, and UQI to 0.833. Guided Loss plays a crucial role in preserving structural information, which is essential for dehazing. In images affected by atmospheric interference, the model must remove haze while maintaining original edge structures. The inclusion of Guided Loss enhances structural consistency, preventing excessive smoothing and preserving fine details of the original image. Notably, the improvement is more pronounced in high-contrast regions and edge areas, highlighting its ability to retain sharp boundaries. Compared to the L2-only model, the addition of Guided Loss results in a more detailed and accurate restoration when tested on the RICE dataset.

When Guided loss was applied, the model achieved PSNR of 32.61 dB, SSIM of 0.966, and UQI of 0.827, further enhancing performance. Watershed loss is particularly effective in preserving edge and contour information, as it guides the model to distinguish between objects and backgrounds within the image. Compared to the baseline L2 Loss model, the Watershed loss model demonstrates superior boundary preservation while effectively removing atmospheric interference. Without Watershed loss, some areas tend to exhibit blurry boundaries, whereas its application ensures that the original contour lines remain sharp. This characteristic is especially beneficial in satellite imagery restoration using the RICE dataset, where clear delineation of buildings, roads, and rivers iscrucial.

The highest performance was achieved when all three loss functions were applied together, resulting in a PSNR of 33.24 dB, SSIM of 0.967, UQI of 0.835. This result suggests that Guided Loss and Watershed loss complement each other, and when combined with L2 Loss, the overall restoration performance is maximized. Guided Loss helps retain global structural information, Watershed loss preserves local edge details, and L2 Loss ensures fundamental pixel-wise restoration. Thus, incorporating all three loss functions achieves the most balanced performance, reinforcing the importance of preserving not only pixel accuracy but also structural integrity and boundary clarity in image restoration.

Dehazing is not merely a pixel restoration problem but also a structural preservation challenge. Even if a model achieves high PSNR and SSIM scores, the restored images may still suffer from blurry edges or unnatural artifacts, making them visually unappealing. To overcome this issue, the proposed loss functions are designed to ensure structural consistency and boundary preservation during dehazing. Guided Loss prevents structural distortions, Watershed loss enhances edge clarity, and L2 Loss maintains overall pixel accuracy. This approach not only minimizes pixel differences but also improves overall restoration quality, making it an optimal solution for realistic image reconstruction, particularly for satellite imagery dehazing in the RICE dataset.

To visually demonstrate the effectiveness of the proposed combination of loss functions in preserving structural information, we applied the Canny edge detection algorithm (with threshold values set to 100 and 200) to the restored images in order to generate edge maps. The comparison targets two types of restoration results: one trained using only the L2 loss, and the other trained with the proposed method incorporating L2 loss, guided loss, and watershed loss. The results are presented in Figure 7, where the left image shows the edge map of the model trained solely with L2 loss, and the right image displays the edge map generated by the proposed model using all loss components. Notably, in the red-highlighted regions of the figure, a clear distinction in structural preservation can be observed between the two methods. While the L2-only model tends to generate broken or blurred edges, the proposed method demonstrates sharper and more continuous edge structures. This visual comparison confirms that the proposed loss functions significantly contribute to learning boundary-aware structural representations. Specifically, the watershed loss introduces spatial constraints based on boundary segmentation, allowing the model to more accurately reconstruct object shapes, especially in regions rich in high-frequency components.

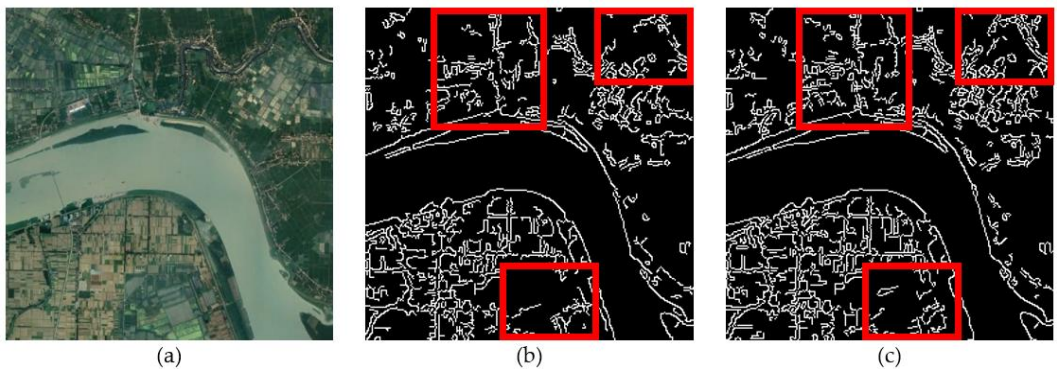

Fig. 7. Original Image (a), Canny Edge Maps: L2 Loss Only (b) vs. Proposed Method with All Loss Functions (c)

Consequently, this edge-based visualization comparison clearly demonstrates the superiority of the proposed method in terms of structural preservation over models trained with single loss functions. It validates that the proposed model not only improves quantitative performance but also achieves greater structural consistency and enhanced perceptual quality.

To quantitatively evaluate the effectiveness of the SwinRRDB module, which serves as a core component of the proposed network, we conducted an ablation study comparing model performance with and without this module. The experiments were carried out on the RICE and StateHaze1K datasets under three haze density conditions: Thin, Moderate, and Thick. Performance was assessed using two widely adopted image restoration metrics: PSNR and SSIM. The results are presented in Table 4.

Table 4. Performance Comparison Based on the Use of SwinRRDB.

|  | SwinRRDB | PSNR | SSIM |
|---|---|---|---|
| RICE | X | 30.50 | 0.952 |
|  | O | **33.24** | **0.967** |
| SateHaze1K (Thin) | X | 22.00 | 0.935 |
|  | O | **24.19** | **0.949** |
| SateHaze1K (Moderate) | X | 25.72 | 0.943 |
|  | O | **28.15** | **0.950** |
| SateHaze1K (Thick) | X | 20.98 | **0.911** |
|  | O | **22.33** | 0.910 |

For the RICE dataset, the model without the SwinRRDB module achieved a PSNR of 30.50 dB and an SSIM of 0.952. When the SwinRRDB module was integrated, the PSNR increased to 33.24 dB and the SSIM improved to 0.967, corresponding to absolute gains of 2.74 dB and 0.015, respectively. These results demonstrate that SwinRRDB contributes significantly to both pixel-level reconstruction accuracy and structural fidelity. Under the StateHaze1K (Thin) haze condition, the model without SwinRRDB achieved a PSNR of 22.00 dB and an SSIM of 0.935. With SwinRRDB, the PSNR improved to 24.19 dB and the SSIM to 0.949, showing enhancements of 2.19 dB and 0.014, respectively. This suggests that the module is effective in recovering fine texture information even under light haze scenarios. In the StateHaze1K (Moderate) condition, excluding the SwinRRDB resulted in a PSNR of 25.72 dB and an SSIM of 0.943. Incorporating the module increased the PSNR to 28.15 dB and SSIM to 0.950, yielding improvements of 2.43 dB and 0.007, indicating its strong capability to preserve structural consistency under medium haze density. For the StateHaze1K (Thick) case, the model without SwinRRDB achieved the lowest performance with a PSNR of 20.98 dB and an SSIM of 0.911. After adding SwinRRDB, the PSNR increased to 22.33 dB, representing a gain of 1.35 dB, while the SSIM slightly decreased to 0.910, showing a marginal drop of 0.001. This implies that in cases of extremely dense haze, where structural information is severely degraded, improvements in perceptual quality (SSIM) may be limited, although pixel-level accuracy still benefits from the module.

Overall, the SwinRRDB module consistently demonstrated performance improvements in both PSNR and SSIM under most conditions, with particularly notable effects observed in low to moderate haze scenarios. This can be attributed to the ability of the proposed SwinRRDB architecture to effectively preserve both local and global information during the dehazing process. Therefore, it can be concluded that SwinRRDB serves as a key performance-enhancing component within the proposed network and plays a crucial role in supporting the overall superiority of the architecture.

## 5. Conclusion

In this study, we propose SUFERNOBWA, a Swin Transformer-based hybrid U-Net network, to effectively remove atmospheric interference in satellite imagery. The proposed model is characterized by global contextual learning using SwinRRDB and structural restoration enhancement based on Watershed loss. To address the challenges of spatial distortion and boundary degradation faced by conventional dehazing techniques, this study introduces a Watershed-based loss function to accurately restore object boundaries. Through this approach, we demonstrate that it is possible to maintain the structural consistency of satellite images while effectively eliminating atmospheric interference.

Experimental results show that the proposed method achieves higher PSNR and SSIM performance compared to state-of-the-art methods on the RICE and SateHaze1K datasets, with a particular advantage in structural restoration. The ablation study confirms that Guided Loss and Watershed loss contribute to improved dehazing performance, and in particular, the Watershed-based loss function plays a crucial role in enhancing object boundary preservation, significantly improving dehazing performance. This suggests that the proposed approach effectively preserves the structural details of key objects in satellite imagery, such as roads, buildings, and terrain contours, while simultaneously removing atmospheric interference.

However, this study has certain limitations. First, color reconstruction errors may occur in high-reflectance areas (e.g., rooftops and water surfaces), which is associated with color inconsistency problems in non-homogeneous haze environments. Second, this study focuses on single-image dehazing, but utilizing multi-temporal data and multi-spectral information could potentially provide more stable restoration performance. Therefore, future research should focus on addressing these challenges to develop a more generalized satellite image restoration model.

In addition, the experiments and evaluations conducted in this study were based on synthetic dataset, and direct validation on real-world satellite imagery was not performed. While synthetic data allows for precise control of atmospheric interference conditions and is advantageous for quantitative evaluation, it does not fully capture the complexity of real environments, such as varying weather conditions, lighting changes, and diverse surface reflectance properties. Therefore, to verify whether the proposed model can deliver comparable dehazing performance in real-world scenarios, additional experiments using real satellite images are essential. Future work should focus on training and evaluating the model with real-world data to rigorously assess its generalizability and practical applicability.

## References

[1] Ghamisi, P.; Rasti, B.; Yokoya, N.; Wang, Q.; Hofle, B.; Bruzzone, L.; Benediktsson, J. A. Multisource and multitemporal data fusion in remote sensing: A comprehensive review of the state of the art. IEEE Geoscience and Remote Sensing Magazine, 2019. 7, 6–39.

[2] Schmitt, M.; Zhu, X. X. Data fusion and remote sensing: An ever-growing relationship. IEEE Geoscience and Remote Sensing Magazine, 2016. 4, 6–23.

[3] Wulder, M. A.; Loveland, T. R.; Roy, D. P.; Crawford, C. J.; Masek, J. G.; Woodcock, C. E.; Zhu, Z. Current status of Landsat program, science, and applications. Remote Sensing of Environment, 2019. 225, 127–147.

[4] Si, J.; Kim, S. GAN-Based Map Generation Technique of Aerial Image Using Residual Blocks and Canny Edge Detector. Applied Sciences, 2024. 14, 10963.

[5] Khan, M. J.; Khan, H. S.; Yousaf, A.; Khurshid, K.; Abbas, A. Modern trends in hyperspectral image analysis: A review. IEEE Access, 2018. 6, 14118–14129.

[6] Yokoya, N.; Grohnfeldt, C.; Chanussot, J. Hyperspectral and multispectral data fusion: A comparative review of the recent literature. IEEE Geoscience and Remote Sensing Magazine, 2017. 5, 29–56.

[7] Cheng, G.; Han, J.; Lu, X. Remote sensing image scene classification: Benchmark and state of the art. Proceedings of the IEEE, 2017. 105, 1865–1883.

[8] Jiang, H.; Peng, M.; Zhong, Y.; Xie, H.; Hao, Z.; Lin, J.; Hu, X. A survey on deep learning-based change detection from high-resolution remote sensing images. Remote Sensing, 2022. 14, 1552.

[9] Sahu, G.; Seal, A.; Bhattacharjee, D.; Nasipuri, M.; Brida, P.; Krejcar, O. Trends and prospects of techniques for haze removal from degraded images: A survey. IEEE Transactions on Emerging Topics in Computational Intelligence, 2022. 6, 762–782.

[10] Liu, J.; Wang, S.; Wang, X.; Ju, M.; Zhang, D. A review of remote sensing image dehazing. Sensors, 2021. 21, 3926.

[11] Ren, W.; Liu, S.; Zhang, H.; Pan, J.; Cao, X.; Yang, M. H. Single image dehazing via multi-scale convolutional neural networks. European Conference on Computer Vision, Netherlands, 11 October–14 October 2016. 154–169.

[12] Hadjimitsis, D. G.; Clayton, C. R. I.; Hope, V. S. An assessment of the effectiveness of atmospheric correction algorithms through the remote sensing of some reservoirs. International Journal of Remote Sensing, 2004. 25. 3651–3674.

[13] Gao, B. C.; Montes, M. J.; Davis, C. O.; Goetz, A. F. Atmospheric correction algorithms for hyperspectral remote sensing data of land and ocean. Remote Sensing of Environment, 2009. 113, 17–24.

[14] Ju, M.; Ding, C.; Ren, W.; Yang, Y.; Zhang, D.; Guo, Y. J. IDE: Image dehazing and exposure using an enhanced atmospheric scattering model. IEEE Transactions on Image Processing, 2021. 30, 2180–2192.

[15] Sharma, T.; Shah, T.; Verma, N. K.; Vasikarla, S. A review on image dehazing algorithms for vision-based applications in outdoor environments. IEEE Applied Imagery Pattern Recognition Workshop, Washington DC, USA, 13 October–15 October 2020. 1–13.

[16] Tseng, C. H.; Chen, L. C.; Wu, J. H.; Lin, F. P.; Sheu, R. K. An automated image dehazing method for flood detection to improve flood alert monitoring systems. Journal of the National Science Foundation of Sri Lanka, 2018. 46.

[17] Zhang, J.; Wang, X.; Yang, C.; Zhang, J.; He, D.; Song, H. Image dehazing based on dark channel prior and brightness enhancement for agricultural remote sensing images from consumer-grade cameras. Computers and Electronics in Agriculture, 2018. 151, 196–206.

[18] Wang, X.; Yang, C.; Zhang, J.; Song, H. Image dehazing based on dark channel prior and brightness enhancement for agricultural monitoring. International Journal of Agricultural and Biological Engineering, 2018. 11, 170–176.

[19] Ni, W.; Gao, X.; Wang, Y. Single satellite image dehazing via linear intensity transformation and local property analysis. Neurocomputing, 2016. 175, 25–39.

[20] Shen, H.; Ding, H.; Zhang, Y.; Cong, X.; Zhao, Z. Q.; Jiang, X. Spatial-frequency adaptive remote sensing image dehazing with mixture of experts. IEEE Transactions on Geoscience and Remote Sensing, 2024. 62, 1-14.

[21] Gui, J.; Cong, X.; Cao, Y.; Ren, W.; Zhang, J.; Zhang, J.; Cao, J.; Tao, D. A comprehensive survey and taxonomy on single image dehazing based on deep learning. ACM Computing Surveys. 2023. 55, 1-37.

[22] Huang, B.; Zhi, L.; Yang, C.; Sun, F.; Song, Y. Single satellite optical imagery dehazing using SAR image prior based on conditional generative adversarial networks. Proceedings of the IEEE/CVF Winter Conference on Applications of Computer Vision Snowmass, CO, USA, 1 March–5 March 2020. 1806–1813.

[23] Chen, X.; Li, Y.; Dai, L.; Kong, C. Hybrid high-resolution learning for single remote sensing satellite image dehazing. IEEE Geoscience and Remote Sensing Letters, 2021. 19, 1–5.

[24] Song, T.; Fan, S.; Li, P.; Jin, J.; Jin, G.; Fan, L. Learning an effective transformer for remote sensing satellite image dehazing. IEEE Geoscience and Remote Sensing Letters, 2023, 20, 1–5.

[25] Yang, L.; Cao, J.; Chen, W.; Wang, H.; He, L. An efficient multi-scale transformer for satellite image dehazing. Expert Systems, 2024. 41, e13575.

[26] Hwang, J.; Yu, C.; Shin, Y. SAR-to-optical image translation using SSIM and perceptual loss based cycle-consistent GAN. International Conference on Information and Communication Technology Convergence, Jeju, Korea, 21 October–23 October 2020. 191–194.

[27] Singh, P.; Komodakis, N. Cloud-GAN: Cloud removal for Sentinel-2 imagery using cyclic consistent generative adversarial networks. IEEE International Geoscience and Remote Sensing Symposium, Valencia, Spain, 22 July–27 July 2018. 1772–1775.

[28] Cai, B.; Xu, X.; Jia, K.; Qing, C.; Tao, D. DehazeNet: An end-to-end system for single image haze removal. IEEE Transactions on Image Processing, 2016. 25, 5187–5198.

[29] Li, S.; Zhou, Y.; Xiang, W. M2SCN: Multi-model self-correcting network for satellite remote sensing single-image dehazing. IEEE Geoscience and Remote Sensing Letters, 2022. 20, 1–5.

[30] Vishwakarma, S.; Punj, D. A Novel Framework for Satellite Image Dehazing Using Advanced Computational Techniques. IEEE Asian Conference on Innovation in Technology (ASIANCON), 2024, 1–7.

[31] He, Y.; Li, C.; Li, X. Remote sensing image dehazing using heterogeneous atmospheric light prior. IEEE Access, 2023. 11, 18805–18820.

[32] Wang, M.; Song, Y.; Wei, P.; Xian, X.; Shi, Y.; Lin, L. IDF-CR: Iterative diffusion process for divide-and-conquer cloud removal in remote-sensing images. IEEE Transactions on Geoscience and Remote Sensing, 2024.

[33] Guo, Y.; Gao, Y.; Liu, W.; Lu, Y.; Qu, J.; He, S.; Ren, W. SCANet: Self-paced semi-curricular attention network for non-homogeneous image dehazing. Proceedings of the IEEE/CVF Conference on Computer Vision and Pattern Recognition, Vancouver, Canada, 18 June–22 June 2023. 1885–1894.

[34] Pan, H. Cloud removal for remote sensing imagery via spatial attention generative adversarial network. arXiv preprint arXiv:2009.13015, 2020.

[35] Song, Y.; He, Z.; Qian, H.; Du, X. Vision transformers for single image dehazing. IEEE Transactions on Image Processing, 2023. 32, 1927-1941.

[36] Lihe, Z.; He, J.; Yuan, Q.; Jin, X.; Xiao, Y.; Zhang, L; Phdnet: A novel physic-aware dehazing network for remote sensing images. Information Fusion. 2024. 106, 102277.

[37] Yang, L.; Cao, J.; Bian, H.; Qu, R.; Guo, H.; Ning, H. Remote Sensing Image Dehazing via Dual-View Knowledge Transfer. Applied Sciences. 2024. 14, 8633.

[38] Liu, Z., Lin, Y., Cao, Y., Hu, H., Wei, Y., Zhang, Z., Lin, S.; Guo, B. Swin transformer: Hierarchical vision transformer using shifted windows. Proceedings of the IEEE/CVF international conference on computer vision, Virtual, 11 October–17 October. 2021. 10012-10022.

[39] Li, C.; Yu, H.; Zhou, S.; Liu, Z.; Guo, Y.; Yin, X.; Zhang, W. Efficient dehazing method for outdoor and remote sensing images. IEEE Journal of Selected Topics in Applied Earth Observations and Remote Sensing, 2023, 16, 4516-4528.

[40] Li, C.; Zhou, S.; Wu, T.; Shi, J.; Guo, F. A Dehazing Method for UAV Remote Sensing Based on Global and Local Feature Collaboration. Remote Sens. 2025, 17, 1688.

[41] Sui, T.; Xiang, G.; Chen, F.; Li, Y.; Tao, X.; Zhou, J.; Hong, J.; Qiu, Z. U-Shaped Dual Attention Vision Mamba Network for Satellite Remote Sensing Single-Image Dehazing. Remote Sens. 2025, 17, 1055.

[42] Dong, W.; Wang, C.; Xu, X. ICL-Net: Inverse cognitive learning network for remote sensing image dehazing. IEEE Journal of Selected Topics in Applied Earth Observations and Remote Sensing., 2024, 17, 16180-16191.

[43] He, K.; Sun, J.; Tang, X. Guided image filtering. IEEE transactions on pattern analysis and machine intelligence, 2012. 35, 1397-1409.

[44] Lin, D.; Xu, G.; Wang, X.; Wang, Y.; Sun, X.; Fu, K. A remote sensing image dataset for cloud removal. arXiv preprint arXiv:1901.00600, 2019.

[45] He, K.; Sun, J.; Tang, X. Single image haze removal using dark channel prior. IEEE Trans. Pattern Anal, 2011. 33, 2341–2353.